\documentclass[sigconf]{acmart}

\usepackage{graphicx}
\usepackage{booktabs}
\usepackage{amsmath}
\usepackage{enumitem}
\usepackage{subcaption}
\usepackage{tabularx}
\usepackage{makecell}
\usepackage{array}

\graphicspath{{figures/}}

\title{RepoEviAgent: Scalable Synthesis of Evidence-Grounded Repository-Level Question Answering Tasks}

\author{Aoqi Hu}
\email{huaoqi800@gmail.com}
\affiliation{%
  \institution{National University of Singapore}
  \city{Singapore}
  \country{Singapore}
}

\author{Yueao Xing}
\email{sd513@bupt.edu.cn}
\affiliation{%
  \institution{Beijing University of Posts and Telecommunications}
  \city{Beijing}
  \country{China}
}

\begin{document}

\begin{abstract}
Real-world software engineering (SWE) tasks require agents to understand repository structures, cross-module dependencies, and runtime behaviors, posing significant challenges for constructing high-quality training and evaluation data. To address this, we propose \textbf{RepoEviAgent}, a scalable SWE data construction framework that integrates quality-aware data curation and role-specialized multi-agent task synthesis. Starting from 8,477 open-source repositories and approximately 2.48 million pull request records, we employ semantic alignment, complexity constraints, and multidimensional quality assessment to identify valuable SWE data. We further develop a sandbox-based multi-agent framework that integrates automated environment construction, execution validation, repository exploration, and evidence-grounded Q\&A synthesis. Based on this framework, we construct \textbf{RepoEviQA}, a benchmark for \textbf{Repo}sitory-Level \textbf{Evi}dence-Grounded \textbf{Q}uestion \textbf{A}nswering, comprising 334 tasks across five programming languages and diverse SWE categories, with rigorous rubric-based evaluation. We evaluate five frontier coding models on RepoEviQA, revealing substantial performance differences across task categories and programming languages.
\end{abstract}

\maketitle

\section{Introduction}
\label{sec:introduction}


Recent advances in Large Language Models (LLMs) have driven code agents beyond standalone code generation toward complex repository-level software engineering (SWE) tasks, including autonomous exploration, issue localization, and code maintenance \citep{jimenez2024swebench, zhang2024autocoderover,wang2025openhands}. Unlike isolated programming problems, real-world SWE tasks inherently involve multi-file project structures, cross-module dependencies, and runtime execution behaviors~\cite{zhang2023repocoder,ding2023crosscodeeval,liu2024repobench}. Consequently, constructing large-scale, high-quality SWE datasets with executable environments has become essential for enabling LLMs to tackle complex repository-level challenges~\cite{li2024deveval}.

\begin{figure}[!t]
    \centering
    \includegraphics[width=\columnwidth]{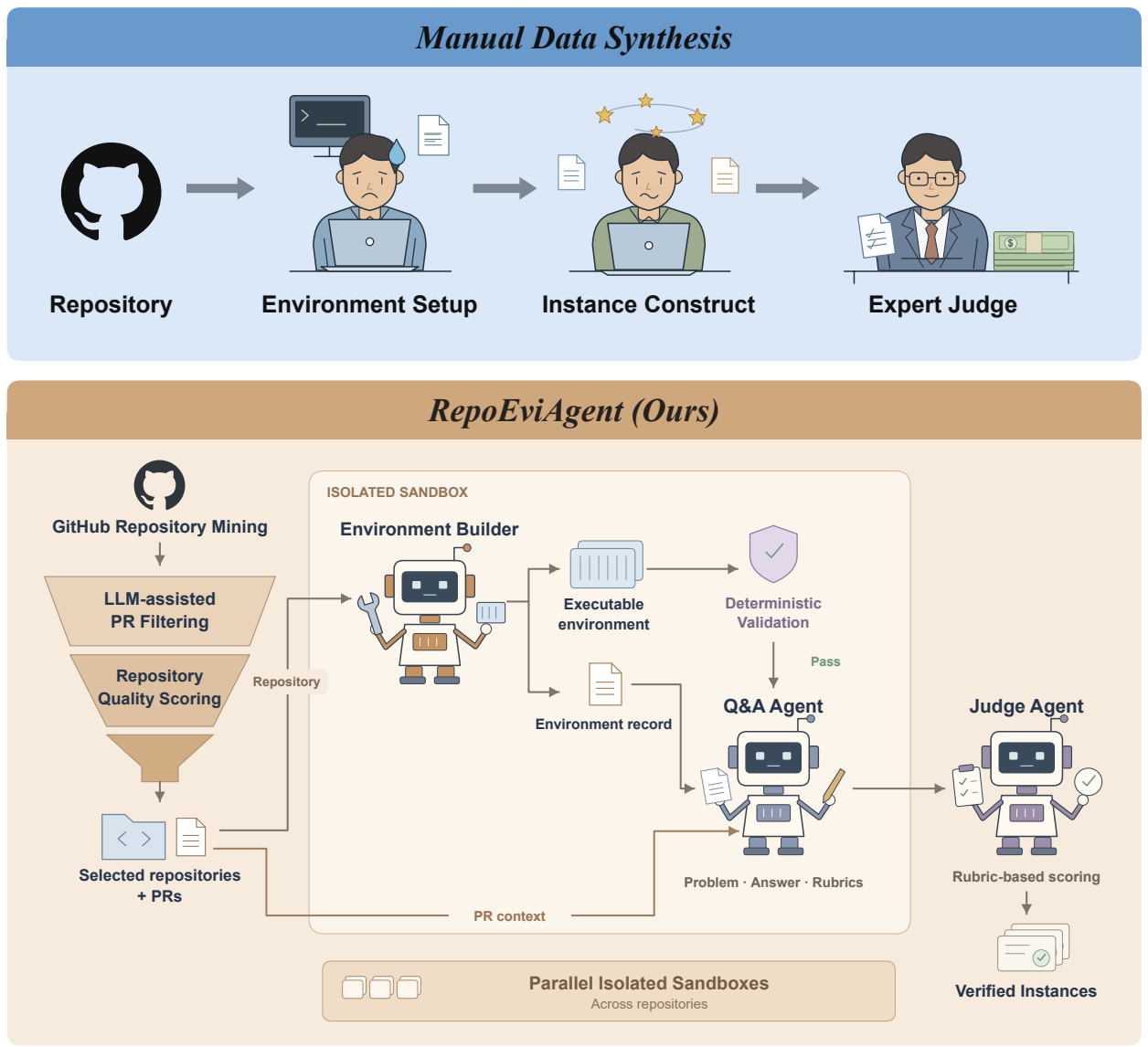}
    \caption{Comparison between manual data synthesis and RepoEviAgent (Ours). Our framework enables scalable and reliable SWE data synthesis through automated environment construction and multi-agent collaboration.}
    \label{fig:difference}
\end{figure}


Despite substantial progress in SWE benchmark construction and training data synthesis \citep{jimenez2024swebench,jain2025r2egym}, existing approaches still face significant challenges in data curation and executable environment construction.  On the one hand, many pull requests (PRs) in real-world repositories involve superficial changes, such as documentation updates or mechanical refactoring, that do not reflect meaningful software behavior changes. Directly converting such records into training or evaluation tasks may introduce low-value samples with insufficient verification evidence. Meanwhile, existing data synthesis methods primarily focus on code generation and issue resolution, lacking systematic mechanisms for synthesizing repository-level  Q\&A tasks. On the other hand, executable environments are essential for ensuring SWE data quality \citep{jain2024r2e,pan2025swegym}, yet existing frameworks remain largely limited to the Python ecosystem \citep{jimenez2024swebench,pan2025swegym,yang2025swesmith, zan2025multiswebench}. Scaling environment construction across programming languages remains challenging, while insufficient validation may introduce workspace contamination and false-positive test results, compromising data reliability~\cite{wang2026swebenchcorrectness}.


To overcome these challenges, we propose a scalable SWE data construction framework for real-world open-source software repositories, integrating quality-aware data curation and multi-agent task synthesis into a unified pipeline, as shown in Figure~\ref{fig:difference}. First, we collect 8,477 highly starred open-source repositories across multiple programming languages and approximately 2.48 million pull request (PR) records from GitHub. We identify candidate data with meaningful software engineering value through LLM-based Issue-PR semantic alignment, structured complexity constraints, and a two-stage multidimensional quality assessment mechanism. For repository-level QA tasks, we design a sandbox-based Role-Specialized Multi-Agent Framework, where specialized agents handle environment construction and evidence-grounded QA synthesis. 


Existing codebase Q\&A benchmarks \citep{liu2021codeqa,liu2024repoqa} primarily focus on static code understanding and long-context semantic retrieval~\cite{wu2024repoformer}, without evaluating agents' ability to verify reasoning conclusions through program execution and dynamic debugging in real execution environments~\cite{gu2024cruxeval}. Data synthesis frameworks for code generation and issue resolution \citep{jain2025r2egym,yang2025swesmith} rely on fail-to-pass tests, which cannot be directly transferred to open-ended, repository-level Q\&A scenarios. To address these limitations, we construct \textbf{RepoEviQA}, comprising 334 tasks across five programming languages and eight SWE task categories, with rigorous rubric-based evaluation to systematically assess code agents' understanding, analysis, and complex reasoning capabilities.


Our main contributions are summarized as follows:

\begin{itemize}[
    label=\textbullet,
    leftmargin=1.5em,
    itemsep=2pt,
    topsep=3pt
]

    \item \textbf{Quality-Aware SWE Data Curation.}
    We propose a quality-aware data curation method
    to identify valuable SWE data from large-scale
    development records in GitHub repositories,
    supporting automated SWE task construction.

    \item \textbf{Scalable Multi-Agent Data Synthesis.}
    We design a role-specialized multi-agent data
    framework that integrates automated environment
    construction, deterministic validation and Q\&A task generation into a
    unified pipeline.

    \item \textbf{High-Quality SWE Benchmark.}
    We construct a multilingual SWE benchmark from
    high-quality synthesized Codebase-QnA tasks,
    covering diverse task categories, and
    systematically evaluate advanced language models
    using a unified mini-SWE-Agent framework.

\end{itemize}

\section{Related Work}
\label{sec:related_work}


\noindent\textbf{Automated Data Construction.} SWE-bench \citep{jimenez2024swebench} constructs real-world code repair tasks from GitHub issues and pull requests, while R2E \citep{jain2024r2e} transforms repositories into interactive execution environments. SWE-Gym \citep{pan2025swegym} introduces executable training tasks with test-based verification, and R2E-Gym \citep{jain2025r2egym} scales task synthesis through automated test generation and back-translation. SWE-smith \citep{yang2025swesmith} further scales code repair data generation through automated environment construction and code perturbation. More recently, MEnvAgent \citep{guo2026menvagent} extends automated environment construction to multiple programming languages through multi-agent planning, execution, and verification. While these studies advance scalable SWE data construction, they primarily focus on code repair and executable environments rather than evidence-grounded repository-level Q\&A synthesis.


\noindent\textbf{SWE Q\&A Benchmark.} Code question-answering benchmarks have evolved from local code understanding to complex repository-level reasoning. Early studies focused on code semantics and retrieval, including CodeQA \citep{liu2021codeqa} and CoSQA \citep{huang2021cosqa}, while RepoQA \citep{liu2024repoqa} extended evaluation to function retrieval in large codebases. Recent benchmarks further emphasize complex software engineering reasoning. SWE-QA \citep{peng2026sweqa} introduces repository-level tasks derived from real GitHub issues, covering cross-file reasoning and multi-hop dependency analysis, while SWE-QA-Pro \citep{cai2026sweqapro} incorporates executable environments and difficulty calibration to assess agents' autonomous exploration and repository-level reasoning capabilities. However, most existing benchmarks still lack reliable runtime execution feedback and strict evaluation rubrics, making it difficult to systematically assess code agents' complex exploration and reasoning capabilities in software environments.

\section{Scalable SWE Data Construction}
\label{sec:method}

\begin{figure*}[!t]
    \centering
    \includegraphics[width=\textwidth]{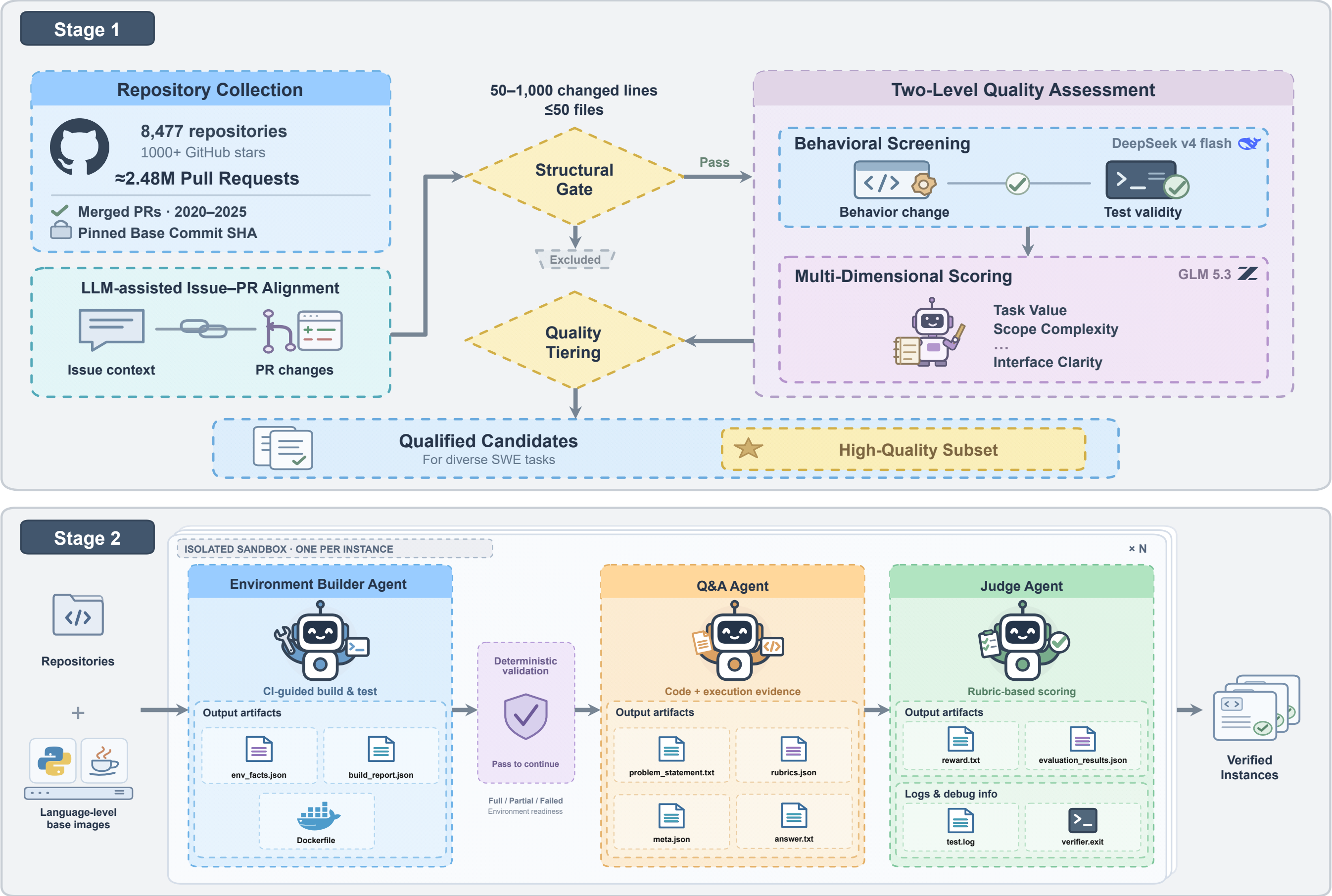}
    \caption{Overview of our automated SWE data synthesis pipeline. Stage 1 performs quality-aware repository sourcing and candidate mining to identify high-quality SWE data. Stage 2 employs our RepoEviAgent framework to synthesize and validate repository-level Q\&A tasks in isolated executable environments.}
    \Description{Description}
    \label{fig:framework}
\end{figure*}

As illustrated in Figure~\ref{fig:framework}, we present our data synthesis framework. The proposed framework builds upon Github repositories, integrating multidimensional quality-aware data curation with a collaborative multi-agent architecture. It enables automated executable environment construction and the synthesis of SWE Q\&A tasks grounded in runtime analysis. Compared with traditional approaches that rely on static rule-based filtering or direct model-based data generation, our framework offers greater flexibility and scalability. In the following sections, we first introduce the strategies for selecting high-quality repositories and candidate data (Section~\ref{sec:repository_sourcing}), then describe the multi-agent data synthesis framework (Section~\ref{sec:multi_agent_synthesis}), and finally explain how high-quality tasks are selected from the synthesized data to construct the \textbf{RepoEviQA}.

\subsection{Framework Overview}

\noindent\textbf{Problem Formulation.} The collection of repositories is formulated as $\mathcal{R}=\{r_1,r_2,\ldots,r_m\}$. Our goal is to develop an automated SWE data synthesis framework $\mathcal{F}$ that generates a high-quality SWE dataset $\mathcal{D}=\mathcal{F}(\mathcal{R})$ with well-defined task descriptions and verifiable evaluation criteria. Each SWE data instance is defined as:
\begin{equation}
d_i=(I_i,p_i,\mathcal{U}_i),
\end{equation}
where $I_i$ denotes the Docker image used to instantiate the task execution environment, $p_i$ represents the problem statement, and $\mathcal{U}_i$ denotes the evaluation rubrics.

\noindent\textbf{Framework Architecture.} Our SWE data construction framework consists of two core modules: \textbf{Data Curation and Agentic Data Synthesis}. Data Curation collects and filters open-source repositories to identify candidate data with high software engineering value. Agentic Data Synthesis employs a collaborative multi-agent architecture, where an environment builder agent constructs executable environments, and a Q\&A synthesis agent explores code repositories to generate high-quality question-answering tasks and rigorous evaluation rubrics. An automated verification mechanism further ensures the quality of synthesized data. Finally, the synthesized instances are solved by strong code models, and the resulting solution trajectories are used to construct supervised fine-tuning (SFT) data.

\subsection{Repository Sourcing and Quality-Aware Candidate Mining}
\label{sec:repository_sourcing}

\noindent\textbf{Repository Collection.} To ensure the authenticity and diversity of synthesized data, we construct a collection of candidate repositories from GitHub, covering different programming languages and software projects. We first collect 8,477 candidate repositories with over 1,000 GitHub stars and approximately 2.48 million pull request (PR) records, extracting key metadata including PR titles, descriptions, associated issues, code changes, and commit SHAs. To ensure data validity and version consistency, we retain only successfully merged PRs with merge dates from January 1, 2020, to before January 1, 2026. We further use the base commit SHA to fix the code baseline, ensuring that each candidate task is associated with a specific repository version.

\noindent\textbf{Issue-PR Alignment and Structural Filtering.} A substantial proportion of PRs lack clear connections to their original issue reports. To address this limitation, we first use an LLM to identify associated issues from PR titles and descriptions, retrieving the corresponding problem background, requirements, and contextual information to establish links between issues and code changes. We then introduce structural constraints, restricting functional code changes to 50--1,000 lines and no more than 50 files, while requiring a non-empty test patch to ensure appropriate task complexity and test evidence. Additionally, we use an LLM to assess whether the gold patch introduces meaningful changes in software behavior and whether the test patch effectively verifies these changes, retaining tasks with practical engineering value while filtering out low-value changes such as documentation updates, dependency upgrades, CI/CD configuration changes, and mechanical refactoring.

\noindent\textbf{Multi-Dimensional Quality Assessment.} Building on structural filtering, we further design a two-stage LLM-based quality assessment mechanism to evaluate candidate tasks at a fine-grained level. First, we use DeepSeek V4 Flash to assess the behavioral relevance of code changes and their relationship to test verification, retaining candidate tasks with meaningful behavioral changes and valid test evidence. We then employ GLM 5.3 to quantitatively evaluate tasks across five dimensions: task value, behavior clarity, test signal, scope complexity, and interface clarity. Based on these multidimensional scores, we adopt a tiered quality filtering strategy that divides candidate tasks into a set meeting basic quality requirements and a high-quality set passing stricter quality criteria.

\subsection{Multi-Agent Codebase-QnA Synthesis}
\label{sec:multi_agent_synthesis}

\noindent\textbf{Codebase Q\&A.} This task aims to evaluate agents' in-depth understanding of codebases across multiple dimensions before making code modifications. To complete these tasks, agents must autonomously explore repositories and analyze cross-module dependencies and function call relationships. When necessary, agents must also configure execution environments and run programs to verify their reasoning, rather than relying solely on static code reading or keyword-based retrieval. Each task is evaluated using fine-grained rubrics and is considered successfully completed only if all must-have criteria are satisfied. The success condition is formally defined as:
\begin{equation}
\operatorname{Success}(d_i,\hat{a}_i)
=
\prod_{u_j \in \mathcal{U}_i^{\mathrm{must}}}
\mathbb{I}\left[J(\hat{a}_i,u_j)=1\right].
\end{equation}
where $J(\hat{a}_i,u_j)$ denotes the evaluator's judgment of whether the generated answer $\hat{a}_i$ satisfies the evaluation criterion $u_j$.

To enable large-scale automated synthesis of Codebase Q\&A tasks, we develop a sandbox-based Role-Specialized Multi-Agent Framework that coordinates environment construction, validity verification, task generation, and quality assessment through a collaborative pipeline. The framework assigns different stages to specialized agents with clearly defined responsibilities and provides an isolated execution environment for each synthesized instance, enabling parallel task synthesis across repositories and scalable expansion.

\begin{figure*}[!t]
    \centering

    \subcaptionbox{Task Distribution by Category\label{fig:result_a}}{%
        \includegraphics[height=6.3cm]{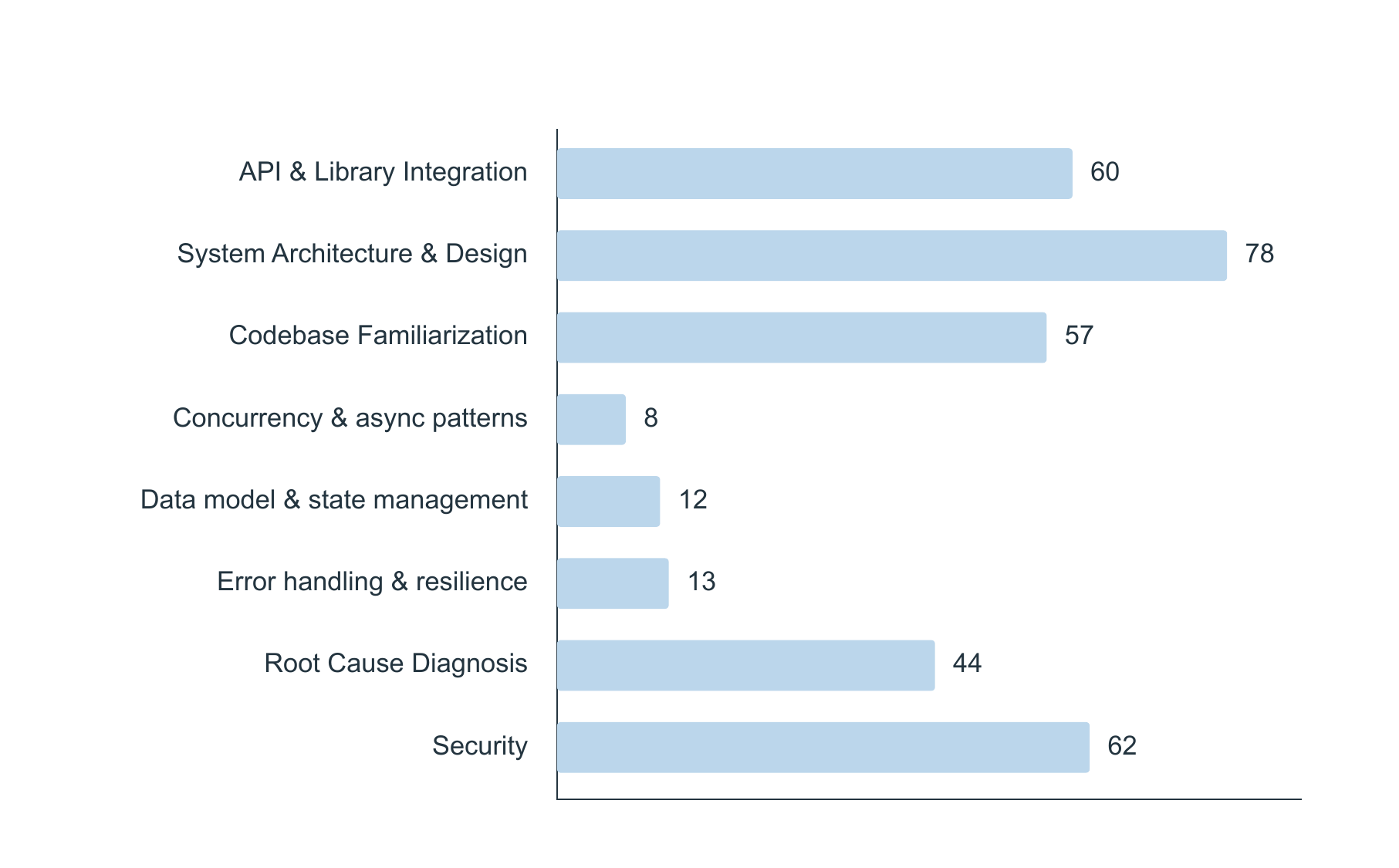}%
    }
    \hfill
    \subcaptionbox{Task Distribution by Language\label{fig:result_b}}{%
        \includegraphics[height=6.3cm]{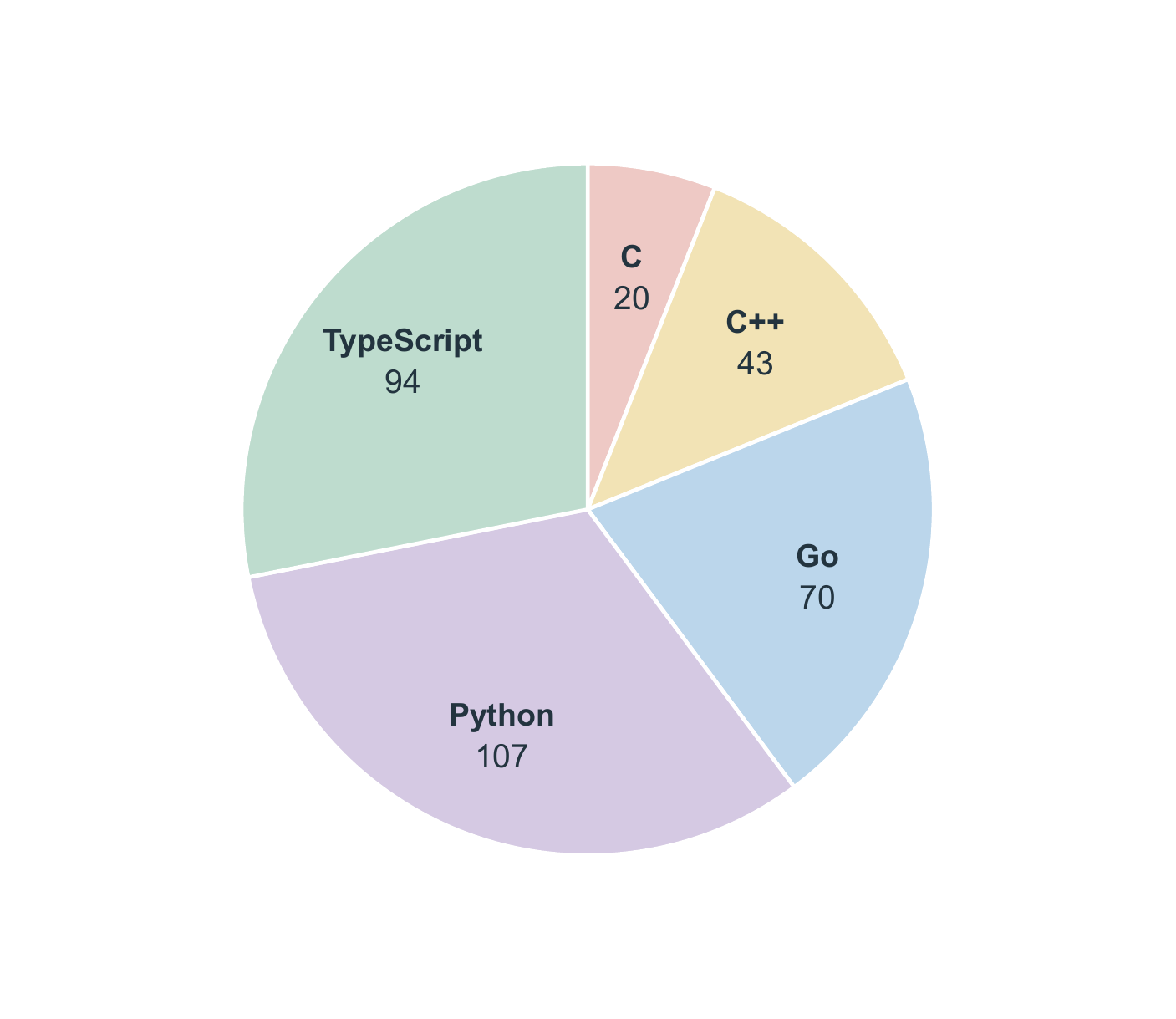}%
    }

    \caption{Task Distribution in \textbf{RepoEviQA}. Distribution of benchmark instances a cross programming languages and software engineering task categories.}
    \label{fig:distribution}
    \Description{Description.}
\end{figure*}

\subsubsection{Role-Specialized Multi-Agent Framework}

The framework assigns an independent sandbox environment to each synthesized instance to ensure isolation during the synthesis process. It consists of two specialized agents, the environment builder agent and the QnA synthesis agent, which enable automated task synthesis across repositories through clearly defined responsibilities and structured information exchange.

\noindent\textbf{Environment Builder Agent.} The Environment Builder Agent is responsible for establishing executable environments for real-world repositories. By analyzing project configurations and dependency requirements, it autonomously sets up the execution environment and validates its functionality through test operations. The agent also produces a structured environment record that captures repository structure, available test commands, and environment configurations for subsequent task synthesis.

\noindent\textbf{QnA Synthesis Agent.} The QnA Synthesis Agent leverages the established execution environment and structured repository information to autonomously explore code, analyze cross-module dependencies, and perform runtime verification when necessary. It generates Q\&A tasks grounded in actual code implementations and execution evidence. Each task includes a problem statement, a reference answer, and structured evaluation rubrics. An LLM-as-a-Judge evaluator automatically scores the reference answers to support subsequent high-quality data selection.

\subsubsection{Environment Construction and Validation}

To accommodate differences in programming languages, toolchains, and dependencies across repositories, we adopt an environment construction strategy based on language-level base images. We prepare base images with the required toolchains for Python, Go, TypeScript, C and C++, avoiding the need to maintain separate base images for individual repositories. Combining language-level base images with task-level isolation improves the scalability and consistency of environment construction across repositories. This process is formally defined as:
\begin{equation}
S_i = \delta\left(B_{\ell(r_i)}, \mathcal{P}_i, r_i\right),
\end{equation}
where $B_{\ell(r_i)}$ denotes the base image corresponding to the programming language of repository $r_i$, $\mathcal{P}_i$ represents the sequence of configuration commands generated by the environment builder agent, $\delta$ denotes the environment state transition function, and $S_i$ represents the resulting repository execution environment.

In practice, the environment builder agent operates within a Docker container initialized with the cloned repository. It first explores the repository structure and analyzes its configuration files. Guided by CI configurations and project dependency specifications, the agent autonomously configures the execution environment and validates its availability through test operations. Based on the verification results, we classify environments into three levels: \textbf{Full}, indicating successful execution of real repository tests; \textbf{Partial}, indicating successful basic builds or module imports; and \textbf{Failed}, indicating an unusable environment. This classification guides subsequent task generation strategies according to the available execution capabilities. To reduce the computational overhead of repeating the entire setup process from base images, the agent records the commands used to install necessary dependencies, enabling reproducible image construction.

To ensure the reliability of automatically constructed environments, we introduce an independent deterministic environment validation mechanism. The validation module checks repository integrity, consistency between the current commit and the target version, and whether the setup process has modified any version-controlled files. For environments classified as \textbf{Full}, the module re-executes the repository tests reported as successful by the agent to confirm their validity. Only instances that meet the predefined environment readiness requirements and pass all validation checks proceed to the subsequent QnA task synthesis stage, ensuring that task generation is grounded in verified repository environments.

\subsubsection{Evidence-Grounded QnA Synthesis and Verification}

After environment construction and validation, the QnA Synthesis Agent adaptively generates diverse codebase Q\&A tasks based on the repository's execution capabilities and structured information. We introduce a diversity control strategy based on task category rotation to cover multiple dimensions of software engineering capabilities, preventing the generation of repetitive questions or excessive concentration on specific task categories.

During task synthesis, the QnA Synthesis Agent explores the implementation logic and runtime behavior of target subsystems through autonomous code search, file inspection, call dependency analysis, and runtime execution when necessary. Unlike simple factual questions generated directly from code text, the synthesized questions must be grounded in specific repository contexts and verifiable code evidence, emphasizing cross-module reasoning and understanding of runtime behavior. For tasks requiring dynamic analysis, the agent should execute existing tests or write and run verification code to gather execution evidence supporting the questions and reference answers. Based on the collected code and runtime evidence, the agent generates a problem statement, a reference answer, and structured evaluation rubrics that specify the key facts, implementation details, and reasoning evidence required in a correct response.

To further ensure synthesis quality, we introduce an automated LLM-as-a-Judge verification mechanism that evaluates reference answers against their corresponding structured rubrics. By combining evidence-grounded task generation with fine-grained automated evaluation, the framework provides traceable quality assessment results for subsequent high-quality Q\&A data selection and benchmark construction.

\subsection{Trajectory Distillation and SFT Data Construction}

\noindent\textbf{Trajectory Distillation.} To construct high-quality training data, we perform agent trajectory distillation using instances that have passed verifier validation. We employ GLM-5.3, a high-performing model for coding tasks, as the expert model and Claude Opus 4.6 as the verifier. For each instance, we perform four independent rollouts with a sampling temperature of 0.95 and a maximum of 250 interaction turns. A trajectory is considered valid only if it satisfies all must-have rubrics. These validated trajectories are then used to construct SFT data for improving student model performance.

\noindent\textbf{SFT Data Construct.} To accommodate different student model context windows, we construct SFT datasets with context lengths of 128K and 256K tokens. We determine trajectory lengths using the student model's tokenizer and chat template. For the 256K version, trajectories exceeding the context limit are discarded, while the remaining trajectories are preserved without modification. For the 128K version, we apply trajectory splitting and LLM-based summarization to compress overlength trajectories while preserving their training value.

\section{Benchmark Construction}
\label{benchmark}

\begin{table*}[!t]
    \centering
    \caption{Category-wise and overall pass rates (\%) of different models on \textbf{RepoEviQA}.}
    \label{tab:category_ps}

    \small
    \setlength{\tabcolsep}{2pt}
    \renewcommand{\arraystretch}{1.2}
    \renewcommand{\tabularxcolumn}[1]{m{#1}}

    \begin{tabularx}{\textwidth}{
        @{}
        >{\raggedright\arraybackslash}m{1.05in}
        *{9}{>{\centering\arraybackslash}X}
        @{}
    }
        \toprule
        \textbf{Model}
        & {\scriptsize\bfseries
            \shortstack[c]{API \&\\Library\\Integration}}
        & {\scriptsize\bfseries
            \shortstack[c]{System\\Architecture\\\& Design}}
        & {\scriptsize\bfseries
            \shortstack[c]{Codebase\\Familiari-\\zation}}
        & {\scriptsize\bfseries
            \shortstack[c]{Concurrency\\\& async\\patterns}}
        & {\scriptsize\bfseries
            \shortstack[c]{Data model\\\& state\\management}}
        & {\scriptsize\bfseries
            \shortstack[c]{Error handling\\\& resilience}}
        & {\scriptsize\bfseries
            \shortstack[c]{Root Cause\\Diagnosis}}
        & {\scriptsize\bfseries Security}
        & {\scriptsize\bfseries Overall} \\
        \midrule

        \multicolumn{10}{@{}l}{\small\textit{mini-SWE-Agent}} \\
        \midrule

        GLM5-3
        & 80.0 & 73.1 & 75.4 & 62.5
        & 75.0 & 61.5 & 81.8 & 79.0 & 76.3 \\

        DeepSeek-V4.1-Flash
        & 70.0 & 56.4 & 71.9 & 12.5
        & 50.0 & 38.5 & 75.0 & 61.3 & 62.9 \\

        Gemini-3.8-Flash
        & 33.3 & 16.7 & 28.1 & 0.0
        & 16.7 & 23.1 & 27.3 & 27.4 & 24.9 \\

        GLM-5.3-Flash
        & 61.7 & 52.6 & 64.9 & 0.0
        & 33.3 & 30.8 & 50.0 & 48.4 & 52.4 \\

        GPT-5.5
        & 70.0 & 46.2 & 50.9 & 0.0
        & 33.3 & 7.7 & 52.3 & 59.7 & 51.5 \\

        \bottomrule
    \end{tabularx}
\end{table*}

To overcome the limitations of existing benchmarks and comprehensively assess the deep codebase understanding capabilities of state-of-the-art SWE agents, we introduce \textbf{RepoEviQA}, a rigorously curated benchmark constructed through a systematic quality assurance pipeline to ensure high-quality tasks and broad representativeness.

\subsection{Quality-Based Benchmark Selection and Filtering}

To construct a high-quality and representative codebase Q\&A benchmark, we design a multi-stage data filtering and quality control pipeline. We first collect 8,477 candidate open-source repositories with over 1000 GitHub stars and apply multi-stage quality filtering to obtain a high-quality repository collection. Based on these repositories, \textbf{RepoEviAgent} generates 28,467 Codebase-QnA instances, which are then evaluated by a Verifier Agent to filter out tasks that fail automated quality checks. Finally, we select 334 qualified tasks based on programming language coverage and task category diversity to construct \textbf{RepoEviQA}. All selected tasks undergo manual review to ensure the accuracy of reference answers and the reliability of evaluation rubrics, enabling systematic assessment of code agents' codebase understanding and reasoning capabilities across diverse software engineering scenarios.

\subsection{Benchmark Composition}

\noindent\textbf{Language Diversity.} \textbf{RepoEviQA} comprises 334 high-quality Q\&A tasks from 168 real-world open-source repositories, spanning five programming languages: Python, TypeScript, Go, C++, and C. These tasks cover diverse software engineering scenarios, ranging from application development to system-level programming. Python, TypeScript, and Go account for the majority of tasks, while C/C++ tasks further broaden the benchmark's coverage of different programming language ecosystems and implementation paradigms.


\noindent\textbf{Task Category Coverage.} \textbf{RepoEviQA} covers eight categories of software engineering Q\&A tasks, reflecting real-world scenarios such as analyzing system architectures, diagnosing runtime issues, and investigating security issues. These tasks evaluate agents' codebase understanding and reasoning capabilities across multiple dimensions, including cross-module dependency analysis, runtime behavior analysis, and system design. The detailed distribution of task categories is shown in Figure~\ref{fig:distribution}.

\subsection{Evaluation Metrics}

We adopt a strict evaluation mechanism based on fine-grained rubrics to assess the quality of agent-generated answers. The rubrics consist of two categories: positive high-level items, which specify the correct factual statements and key reasoning evidence that an answer must include, and negative high-level items, which identify incorrect statements that should not appear in an answer. For each generated answer, the Verifier evaluates it against predefined rubrics, and a task is considered successfully completed only if all must-have rubrics are satisfied. To ensure consistency and fairness across different models, we use Claude Opus 4.6 as the unified verification model for all evaluations.

\section{Experiment}
\label{sec:experiment}

\begin{table*}[t]
\centering
\caption{Performance of reference models and our fine-tuned models on SWE-Atlas-QnA.}
\label{tab:sft_performance}

\small
\renewcommand{\arraystretch}{1.2}
\setlength{\tabcolsep}{5pt}

\begin{tabular*}{\textwidth}{@{\extracolsep{\fill}}lc!{\vrule width 0.6pt}lccc@{}}
\toprule

\multicolumn{2}{c!{\vrule width 0.6pt}}{\textit{Panel A. Reference Models (124 Tasks)}}
&
\multicolumn{4}{c}{\textit{Panel B. Our Fine-tuned Models (60 Tasks)}} \\

\midrule

\textbf{Model}
& \textbf{Resolved Rate (\%)}
& \textbf{Model}
& \textbf{Resolved Rate (\%)}
& \textbf{Avg. Score}
& \textbf{Avg. Length (chars)} \\

\midrule

Opus 4.7
& 36.02
& Qwen3.5-27B (Base)
& 18.3 & 0.666 & 7,693 \\

GPT-5.4
& \textbf{36.30}
& Qwen3.5-27B (SFT)
& \textbf{53.3} {\color{green!55!black}(+35.0)}
& \textbf{0.928} & 15,763 \\

GLM-5
& 20.50
& Qwen3.6-35B-A3B (Base)
& 20.0 & 0.690 & 8,955 \\

Kimi K2.5
& 13.10
& Qwen3.6-35B-A3B (SFT)
& \textbf{38.3} {\color{green!55!black}(+18.3)}
& \textbf{0.845} & 18,457 \\

\bottomrule
\end{tabular*}

\end{table*}

\begin{figure*}[!t]
    \centering
    \includegraphics[width=\textwidth]{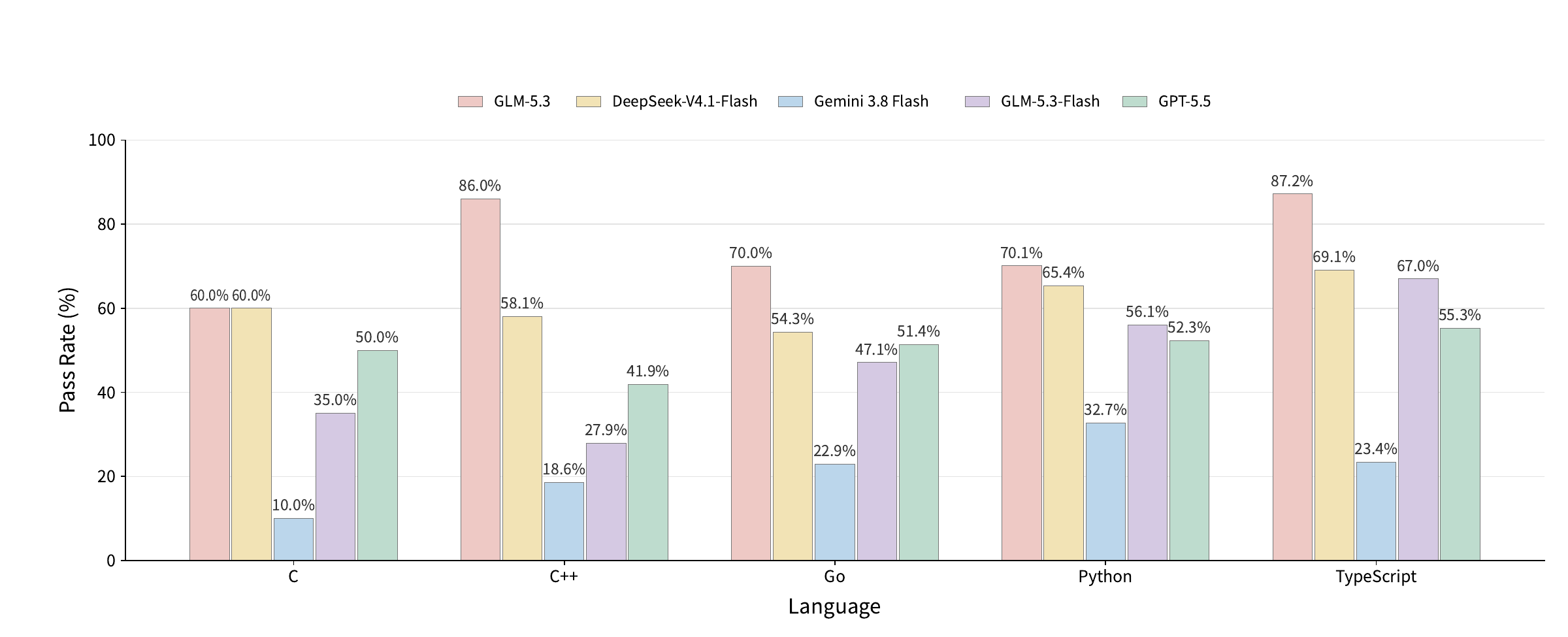}
    \caption{Pass rates of five models across five programming languages in RepoEviQA.}
    \Description{Description}
    \label{fig:language_ps}
\end{figure*}


In this section, we systematically evaluate five representative LLMs on \textbf{RepoEviQA} to assess their codebase understanding and reasoning capabilities. Using a unified agent scaffold, we compare models' overall task completion performance and analyze their task-level performance. We further examine performance differences across software engineering task categories and programming languages. 

\subsection{Experiment Setup}


\noindent\textbf{Agent Scaffolding.} We adopt mini-SWE-agent \citep{yang2024sweagent}, a lightweight open-source SWE agent framework, as the unified execution scaffold for all experiments. Through its streamlined agent architecture and command-line tool interface, the framework enables models to autonomously explore code repositories, execute shell commands, and analyze program execution results in isolated environments. We select mini-SWE-agent for its lightweight design, which uses just the bash command as a tool for the agent, as a minimal harness to evaluate some models under a common scaffold. It reduces the influence of differences in agent frameworks on evaluation results, thereby ensuring experimental reproducibility and fairness.


\noindent\textbf{Model Details.} To comprehensively evaluate the performance of different frontier coding models on real-world repository understanding and complex reasoning tasks, we select five representative advanced models: GPT-5.5 \citep{openai2026gpt55}, DeepSeek-V4.1-Flash \citep{deepseek2026v41flash}, Gemini-3.8-Flash \citep{googledeepmind2026gemini38flash}, GLM-5.3-Flash \citep{zai2026glm53flash}, and GLM-5.3 \citep{zai2026glm53}. These models span different model families and capability profiles, enabling a systematic comparison of software engineering capabilities, including repository exploration, cross-module reasoning, and runtime analysis.


\noindent\textbf{Evaluation Metrics.} We evaluate agent-generated responses using the rubric-based evaluation protocol defined in Section 4.3. Each task contains a set of mandatory evaluation criteria, and a response is considered successful only if it satisfies all must-have rubrics. We use pass rate as the primary evaluation metric, defined as the proportion of successfully completed tasks among all evaluated tasks.

\subsection{Experiment Results}

\newcommand{\pct}[1]{{\normalsize #1\%}}

Table~\ref{tab:category_ps} presents the overall evaluation results of five models on \textbf{RepoEviQA}. The pass rates range from 24.85\% to 76.35\%, with a gap of 51.50 percentage points between the highest and lowest scores. GLM-5.3 achieves the highest overall pass rate of 76.35\%, successfully completing 255 out of 334 evaluation instances, followed by DeepSeek-V4.1-Flash with a pass rate of 62.87\% and 210 completed tasks. GLM-5.3-Flash and GPT-5.5 achieve pass rates of 52.40\% and 51.50\%, respectively, while Gemini-3.8-Flash achieves 24.85\%. Evaluated on the same set of tasks under identical repository exploration and tool interaction conditions, these models exhibit substantial differences in task completion performance. This clear performance separation indicates that \textbf{RepoEviQA} can effectively characterize the performance distribution of current coding agents on repository-level understanding tasks.

Table~\ref{tab:sft_performance} presents the performance of reference models and our fine-tuned models on SWE-Atlas-QnA~\cite{raghavendra2026swe}. Panel A reports the results of reference models evaluated on the full set of 124 tasks, while Panel B presents our fine-tuning results on a subset of 60 tasks. Fine-tuning on our synthesized data yields substantial improvements across both models. Specifically, Qwen3.5-27B achieves a resolved rate of 53.3\%, improving by 35.0 percentage points over its base model, while Qwen3.6-35B-A3B improves from 20.0\% to 38.3\%, a gain of 18.3 percentage points. Both models also exhibit improvements in average scores, indicating enhanced response quality. These results demonstrate the effectiveness of our synthesized training data in improving repository-level question-answering capabilities across different model architectures.
\section{Further Analysis}
\label{sec:futher_analysis}

\subsection{Performance Across Task Categories}

\begin{figure}[!t]
    \centering
    \includegraphics[width=\columnwidth]{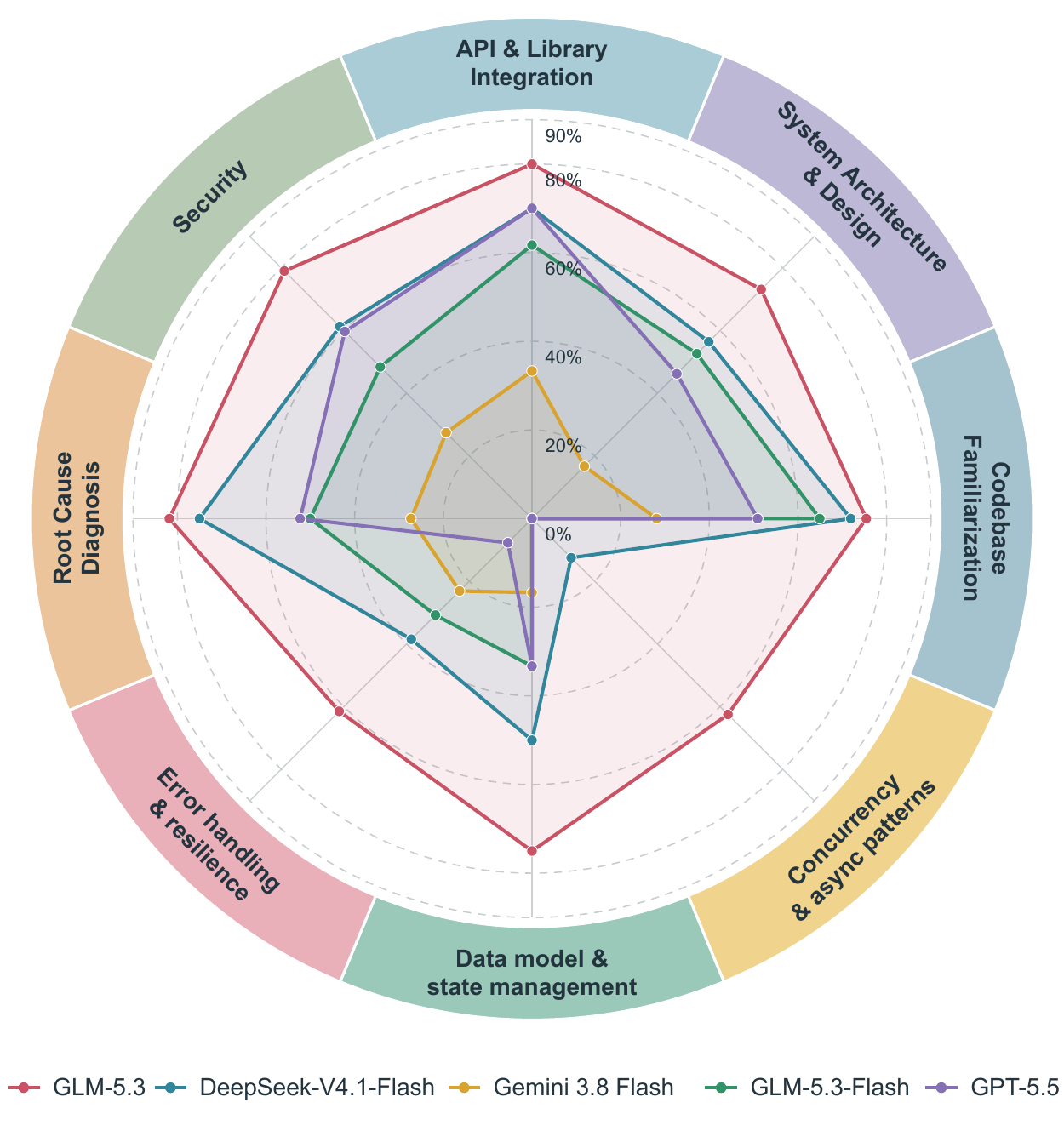}
    \caption{Pass rates of five models across eight software engineering task categories in RepoEviQA.}
    \label{fig:radar}
\end{figure}

Repository-level code understanding involves multiple levels of software engineering capabilities, ranging from understanding local APIs and code behavior to cross-module architectural analysis, root cause diagnosis, and reasoning about complex execution logic. To further analyze differences in model capabilities across specific software engineering scenarios, we report the pass rates of five models across eight task categories in \textbf{RepoEviQA}, as shown in Figure~\ref{fig:radar}.


\noindent\textbf{Overall Performance Across Categories.} Model performance varies considerably across task categories. API \& Library Integration is among the categories with higher overall pass rates, achieving an average pass rate of 63.0\% across the five models. Codebase Familiarization, Root Cause Diagnosis, and Security achieve average pass rates of 58.2\%, 57.3\%, and 55.2\%, respectively. In contrast, the average pass rates for System Architecture \& Design and Data Model \& State Management decrease to 49.0\% and 41.7\%, respectively, while those for Error Handling \& Resilience and Concurrency \& Async Patterns further decline to 32.3\% and 15.0\%. These results indicate that different software engineering tasks pose varying levels of challenges to current coding agents, whose task completion capabilities are not consistent across different scenarios.


\noindent\textbf{Local Code Understanding vs. System-level Reasoning.} In terms of local code understanding, several models perform relatively well on API \& Library Integration and Codebase Familiarization tasks. For example, GLM-5.3 and DeepSeek-V4.1-Flash achieve pass rates of 75.4\% and 71.9\% on Codebase Familiarization, respectively. On API \& Library Integration, GLM-5.3 achieves 80.0\%, while DeepSeek-V4.1-Flash and GPT-5.5 both reach 70.0\%. However, when moving to System Architecture \& Design tasks, the pass rates of all five models decrease by 6.9 to 23.8 percentage points. Specifically, GPT-5.5 drops from 70.0\% to 46.2\%, while DeepSeek-V4.1-Flash declines from 70.0\% to 56.4\%. Unlike API-related tasks that focus on analyzing specific code entities, system architecture tasks typically require integrating implementation details distributed across different modules and establishing an understanding of component dependencies and overall system design. This consistent performance decline indicates that cross-module information integration and system-level reasoning remain important challenges for coding agents.


\noindent\textbf{Diagnostic Reasoning and Security Analysis.} Root cause diagnosis and security analysis further reveal differences in model capabilities for complex software engineering reasoning. In Root Cause Diagnosis, GLM-5.3 (81.8\%) and DeepSeek-V4.1-Flash (75.0\%) substantially outperform GPT-5.5 (52.3\%) and GLM-5.3-Flash (50.0\%). A similar performance gap emerges in Security, where GLM-5.3 leads with a pass rate of 79.0\%, compared with 61.3\% for DeepSeek-V4.1-Flash, 59.7\% for GPT-5.5, and only 27.4\% for Gemini-3.8-Flash. These tasks require models not only to identify relevant code behavior but also to form evidence-based analytical conclusions by considering control flow, module interactions, and potential security risks. The substantial performance differences across models in these two categories indicate that identifying code-level facts does not necessarily imply the ability to perform further engineering reasoning and problem analysis.


\noindent\textbf{Complex Execution Semantics and Robustness.} Models exhibit more pronounced capability limitations on tasks involving complex program behavior. In Data Model \& State Management, GLM-5.3 achieves a pass rate of 75.0\%, while GPT-5.5 and GLM-5.3-Flash both achieve 33.3\%. In Error Handling \& Resilience, GLM-5.3 successfully completes eight tasks, whereas GPT-5.5 completes only one. In Concurrency \& Async Patterns, GLM-5.3 successfully completes five tasks, DeepSeek-V4.1-Flash completes only one, and the remaining three models fail to complete any tasks. These results suggest that tasks involving state transitions, exception handling, and asynchronous execution logic may pose substantial challenges to current models.


\noindent\textbf{Cross-model Capability Profiles.} Category-level results reveal distinct software engineering capability profiles across models. GLM-5.3 achieves the highest pass rate in all eight categories, exceeding 60\% in each. Meanwhile, models with similar overall performance may exhibit different strengths. Despite comparable overall pass rates of 52.4\% and 51.5\%, GLM-5.3-Flash outperforms GPT-5.5 by 14.0 percentage points on Codebase Familiarization and 6.4 points on System Architecture \& Design. Conversely, GPT-5.5 leads by 8.3 and 11.3 percentage points on API \& Library Integration and Security, respectively. These findings demonstrate that \textbf{RepoEviQA} can reveal differences in software engineering capabilities that are not captured by overall pass rates, providing a more fine-grained assessment of coding agents across diverse repository-level tasks.

\begin{figure}[!t]
    \centering
    \includegraphics[width=\columnwidth]{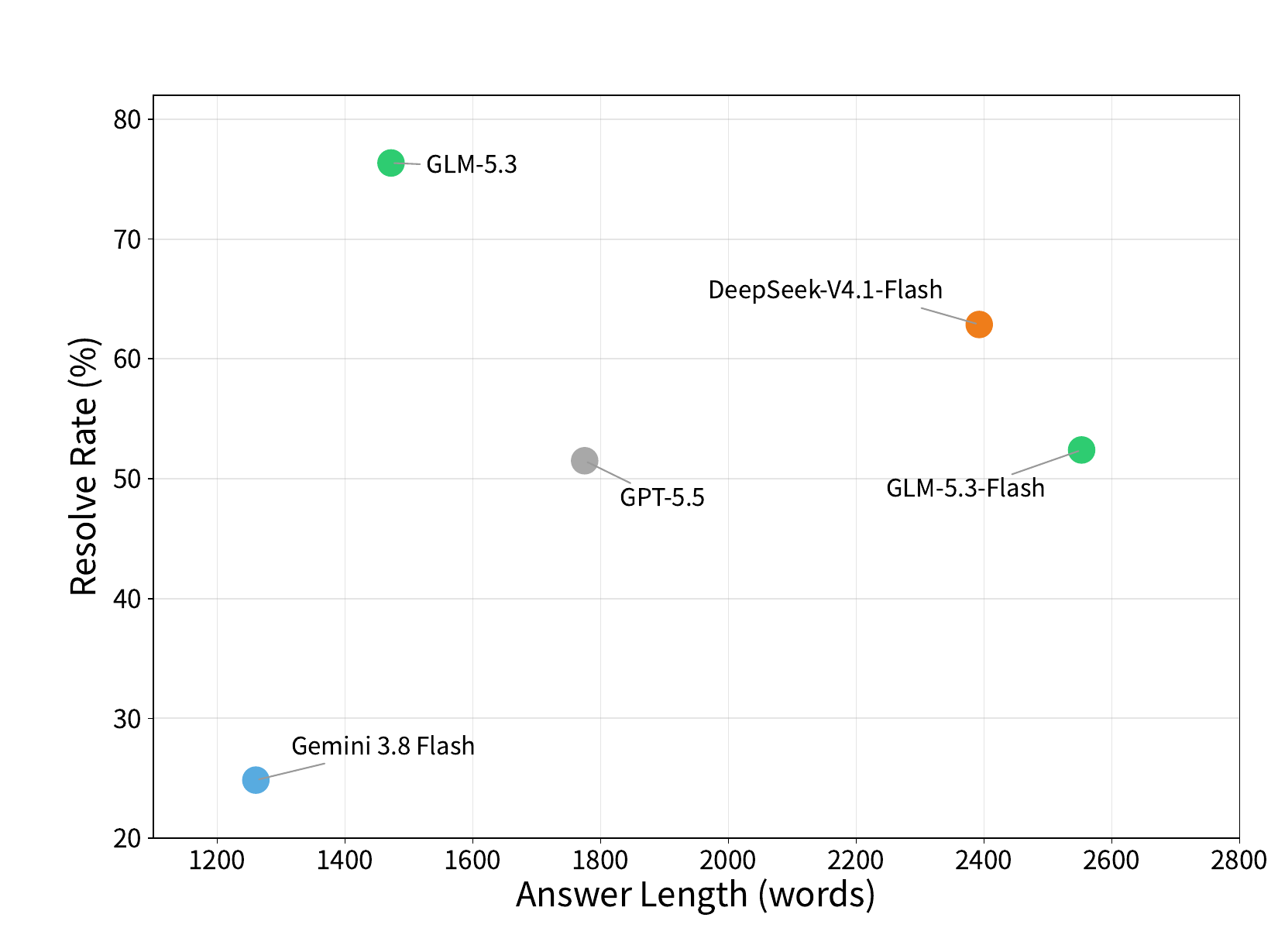}
    \caption{Average final-answer length versus overall pass rate for five models on RepoEviQA. Each point represents one model.}
    \label{fig:length}
\end{figure}

\subsection{Performance Across Programming Languages}


Real-world software repositories written in different programming languages exhibit distinct characteristics in API design, dependency management, and execution mechanisms. To examine repository-level understanding capabilities across heterogeneous software ecosystems, we further compare the performance of five models on tasks involving Python, TypeScript, Go, C++, and C. The distribution of \textbf{RepoEviQA} tasks across these programming languages is shown in Figure~\ref{fig:language_ps}.


\noindent\textbf{Overall Performance Across Languages.} Model performance varies considerably across programming languages. The average pass rates of the five models reach 60.4\% on TypeScript and 55.3\% on Python, compared with 49.1\%, 46.5\%, and 43.0\% on Go, C++, and C, respectively. GLM-5.3 performs particularly well on TypeScript and C++, achieving pass rates of 87.2\% and 86.0\%, approximately 16–17 percentage points higher than its performance on Python and Go. GPT-5.5 exhibits a different performance profile, with pass rates ranging from 50.0\% to 55.3\% across TypeScript, Python, Go, and C, but declining to 41.9\% on C++. These results indicate that performance patterns vary across programming languages, revealing language-specific characteristics in models' repository-level understanding capabilities.


\noindent\textbf{Performance Differences Across Models.} Performance across programming languages further reveals differences in model capabilities. DeepSeek-V4.1-Flash performs relatively well on TypeScript and Python, achieving pass rates of 69.1\% and 65.4\%, with its TypeScript performance exceeding that on Go by 14.8 percentage points. GLM-5.3-Flash also achieves a relatively high pass rate of 67.0\% on TypeScript, trailing DeepSeek-V4.1-Flash by only 2.1 percentage points. However, the gap widens to 9.3 percentage points on Python. In contrast, Gemini-3.8-Flash records pass rates below 33\% across all five languages. Its highest pass rate is 32.7\% on Python, while performance on C and C++ drops to 10.0\% and 18.6\%. These findings suggest that models differ not only in their overall task completion capabilities but also in how their strengths and limitations vary across programming languages.


\noindent\textbf{Performance on System-level Languages.} C and C++ tasks provide an additional evaluation dimension for system-level software engineering scenarios. Repositories written in these languages involve low-level interfaces, resource management, and language-specific implementation mechanisms, requiring models to understand code behavior in the context of concrete program structures. On C++ tasks, GLM-5.3 outperforms GPT-5.5 by 44.1 percentage points, whereas the gap narrows to 10.0 percentage points on C. Meanwhile, DeepSeek-V4.1-Flash achieves comparable pass rates on C and C++, at 60.0\% and 58.1\%. GLM-5.3-Flash records 35.0\% on C and 27.9\% on C++, trailing DeepSeek-V4.1-Flash by 30.2 percentage points on the latter. These results demonstrate that even among tasks involving system-level programming languages, models can exhibit distinct performance patterns, further highlighting the value of language-level evaluation for analyzing repository understanding capabilities.

Overall, task categories and programming languages provide two complementary evaluation dimensions. Category-level results capture differences in model capabilities across specific software engineering tasks, while language-level results characterize how these capabilities manifest across different software ecosystems. Combining these two dimensions allows \textbf{RepoEviQA} to characterize the repository-level understanding capabilities of coding agents from multiple perspectives, providing richer insights into model performance than a single overall pass rate.

\subsection{Answer Length and Task Performance}


To further investigate the relationship between model response behavior and task completion, we calculate the average word count of each model's final answers and compare it with the overall pass rate, as shown in Figure~\ref{fig:length}.


The average answer lengths of the five models range from 1,260 to 2,553 words, but answer length and task pass rate do not exhibit a consistent trend. GLM-5.3 achieves the highest pass rate of 76.35\% with an average answer length of approximately 1,472 words, generating about 920 fewer words than DeepSeek-V4.1-Flash while outperforming it by 13.48 percentage points. A similar difference is observed between GLM-5.3 and GLM-5.3-Flash: the latter produces answers approximately 1.73 times as long as those of the former, yet its pass rate is 23.95 percentage points lower. Meanwhile, GPT-5.5 generates an average of approximately 1,775 words per answer, with a pass rate of 51.50\%. Gemini-3.8-Flash produces the most concise answers, averaging only 1,260 words, but also records the lowest pass rate of 24.85\%.


Across the five models, task pass rates do not consistently increase with average answer length. This observation aligns with the evaluation approach adopted in \textbf{RepoEviQA}. For repository-level code Q\&A, answer quality primarily depends on whether a model can accurately understand relevant implementations, identify the key facts required to address a question, and provide complete, code-grounded analytical conclusions. Rubric-based evaluation assesses factual correctness and reasoning completeness by checking each mandatory evaluation criterion. Answer length analysis complements this observation from the perspective of model output behavior, revealing differences in how coding agents organize their answers and perform on tasks.
\section{Conclusion}
\label{sec:conclusion}

In this paper, we presented \textbf{RepoEviAgent},a scalable framework for automated SWE data synthesis from real-world software repositories. Our framework addresses key challenges in SWE data construction through quality-aware data curation and a role-specialized multi-agent architecture. By integrating automated environment construction, execution validation, and evidence-grounded task synthesis, the framework enables the generation of high-quality SWE data across multiple programming languages. We further leveraged this framework to construct \textbf{RepoEviQA}, a repository-level Q\&A benchmark comprising 334 tasks with rigorous rubric-based evaluation. Experiments on five frontier coding models demonstrate the benchmark's ability to reveal performance differences across diverse software engineering scenarios. Overall, our work provides a systematic approach to scalable and reliable SWE data synthesis, supporting future research on the training and evaluation of coding agents.

\bibliographystyle{ACM-Reference-Format}
\bibliography{references}

\end{document}